\documentclass[letterpaper, 10 pt, conference]{ieeeconf}  

\IEEEoverridecommandlockouts                              

\usepackage[most]{tcolorbox}
\usepackage{xcolor}
\usepackage{cite}
\usepackage{hyperref}
\usepackage{amssymb}
\usepackage{booktabs}
\usepackage{adjustbox}
\usepackage{caption}
\usepackage{amsmath}
\let\labelindent\relax
\usepackage{enumitem}
\usepackage{subcaption} 

\title{\LARGE \bf
The Better You Learn, The Smarter You Prune: Towards Efficient Vision-language-action Models via Differentiable Token Pruning
}

\author{
Titong Jiang$^{1,2}$*, Xuefeng Jiang$^{1,3}$*, Yuan Ma$^{1\dagger}$, Xin Wen$^{1}$, Bailin Li$^{1}$,\\
Kun Zhan$^{1}$, Peng Jia$^{1}$, Yahui Liu$^{2}$, Sheng Sun$^{3}$, and Xianpeng Lang$^{1\ddagger}$%
\thanks{$^{1}$~LiAuto Inc., $^{2}$~School of Vehicle and Mobility, Tsinghua University, $^{3}$~Institute of Computing Technology, Chinese Academy of Sciences}%
\thanks{${*}$~Equal contributions.}%
\thanks{$^{\dagger}$~Project Lead.}%
\thanks{$^{\ddagger}$~Corresponding author. Email: \url{langxianpeng@lixiang.com}}%
}

\begin{document}

\maketitle
\thispagestyle{empty}
\pagestyle{empty}

\begin{abstract}
We present LightVLA, a simple yet effective differentiable token pruning framework for vision-language-action (VLA) models.
While VLA models have shown impressive capability in executing real-world robotic tasks, their deployment on resource-constrained platforms is often bottlenecked by the heavy attention-based computation over large sets of visual tokens.  
LightVLA addresses this challenge through adaptive, performance-driven pruning of visual tokens: It generates dynamic queries to evaluate visual token importance, and adopts Gumbel softmax to enable differentiable token selection. Through fine-tuning, LightVLA learns to preserve the most informative visual tokens while pruning tokens which do not contribute to task execution, thereby improving efficiency and performance simultaneously. Notably, LightVLA requires no heuristic “magic numbers” and introduces no additional trainable parameters, making it compatible with modern inference frameworks.
Experimental results demonstrate that LightVLA outperforms different VLA models and existing token pruning methods across diverse tasks on the LIBERO benchmark, achieving higher success rates with substantially reduced computational overhead.
Specifically, LightVLA reduces FLOPs and latency by 59.1\% and 38.2\% respectively, with a 2.6\% improvement in task success rate.
Meanwhile, we also investigate the learnable query-based token pruning method LightVLA$^*$ with additional trainable parameters, which also achieves satisfactory performance.
Our work reveals that as VLA pursues optimal performance, LightVLA spontaneously learns to prune tokens from a performance-driven perspective.
To the best of our knowledge, LightVLA is the first work to apply adaptive visual token pruning to VLA tasks with the collateral goals of efficiency and performance, marking a significant step toward more efficient, powerful and practical real-time robotic systems. Project site: \href{https://liauto-research.github.io/LightVLA/}{\textit{https://liauto-research.github.io/LightVLA}}.
\end{abstract}

\section{Introduction}

From large-scale industrial operations to personal healthcare and leisure activities, robotics have been reshaping nearly every facet of human life. 
Recently, robotics witnessed the rise of embodied intelligence as its newest technical leap, when artificial intelligence (AI) is introduced into robotics, thanks to the emergence of vision-language-action (VLA) models. 
VLA models can be regarded as a family of large vision-language models (VLMs) which directly translate visual information and language  instructions into executable action policies.
Leveraging the general knowledge and reasoning capabilities inherited from LLMs, VLA models have shown transformative potential in tackling complex robotic reasoning, planning and manipulation tasks\cite{gr-3,openvla,pi0,pifast,pihalf,cogact,oft,molmoact}.

Unfortunately, the success of VLA models also comes with high computational complexity.
As typical VLA models usually include a large language model (LLM) with billion-scale parameters, the expensive attention-based computational cost and high forward latency of VLA models hinder them from real-time applications on systems with computing constraints at edge devices such as household robots \cite{figure2025helix} and autonomous vehicles~\cite{cmt,transdiffuser}. 
As such, the acceleration techniques for VLA models plays a significant role in making VLA models more efficient and practical \cite{efficientvla}.

\begin{figure}[tbp]
    \centering
    \includegraphics[width=\linewidth]{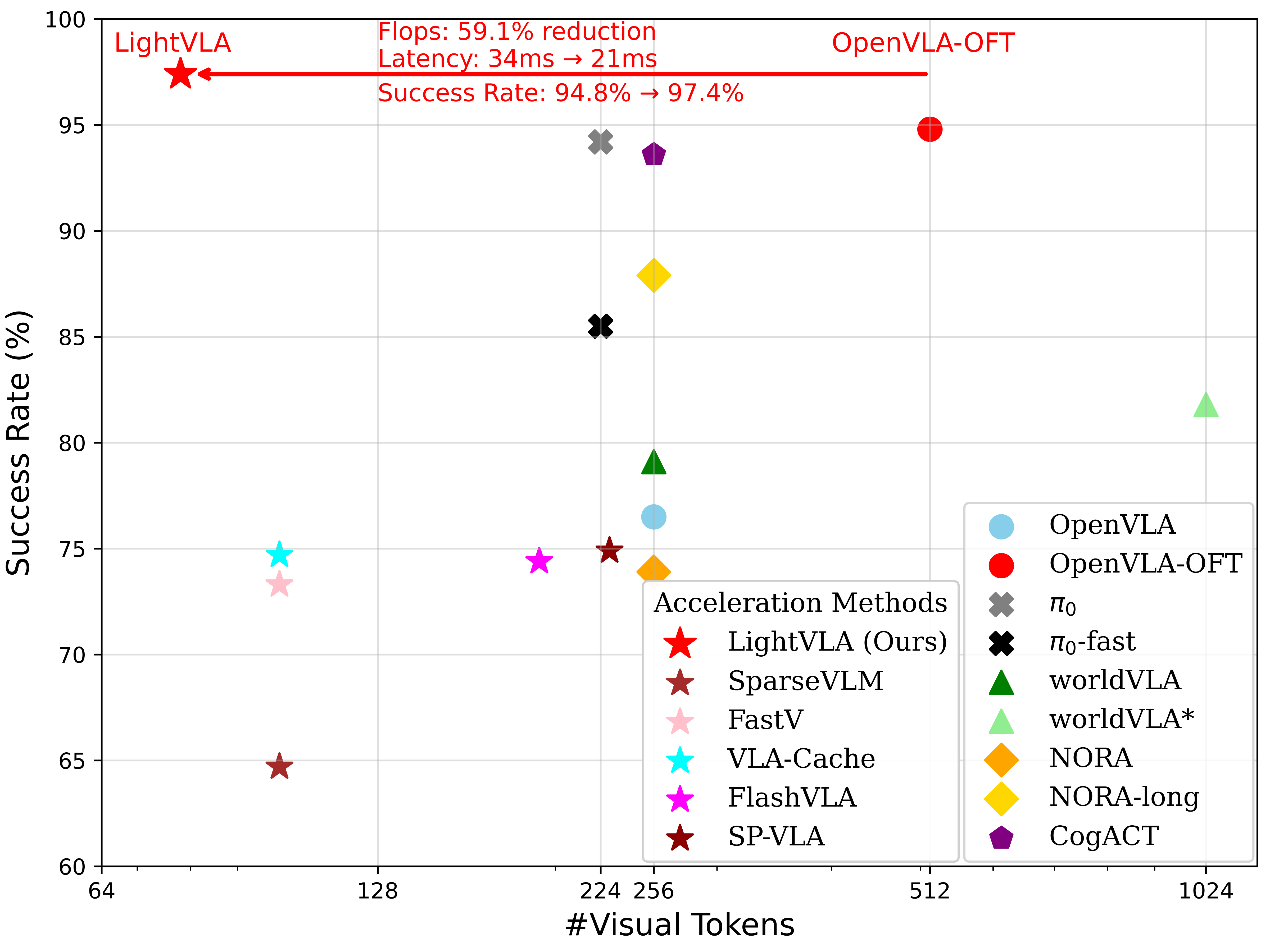}
    \caption{LightVLA achieves better performance than common VLA models and acceleration methods with fewer visual tokens, yielding efficient computation and lower latency.}
    \label{fig:light_vit_1}
\end{figure}

Many acceleration approaches for VLMs and VLA models have been explored in prior studies, including model quantization, layer skipping \cite{efficientvla}, token pruning \cite{llavamini,fastv,sparsevlm,prumerge,pyramiddrop,lvpruning,VisPruner,CDPruner} and lightweight model design \cite{nora,smolvla,tinyvla}.
Among these acceleration approaches, visual token pruning is of particular interest, as the dominant majority of input tokens for VLA models are visual tokens.
Given the sparse nature of vision modality \cite{fastv}, visual tokens often convey little or redundant information, which brings prominent yet unnecessary computational burden for VLA models.
Meanwhile, though visual token pruning has been relatively widely studied in VLMs~\cite{fastv,sparsevlm,llavamini}, recent studies~\cite{flashvla,spvla,vlacache} show that these methods produce unsatisfactory performance when transferring to VLA models, as VLMs focuses on global semantics while specialized robotic tasks depends more on local semantics.
Therefore, visual token pruning oriented for VLA models holds great potential while the related exploration is limited.

It is widely acknowledged that there is a trade-off relationship between efficiency and generalization performance for the VLA acceleration.
Consequently, previous visual token pruning methods often choose efficiency as the priority at the acceptable cost of performance degradation. 
For example, EfficientVLA~\cite{efficientvla} first sets the retained token number as a hyperparameter, and then proposes several approaches to minimize the performance drop caused by token reduction.
However, we argue that efficiency and performance are not intrinsically contradictory.
Note that the sparsity of the visual input is not only contributing to computational inefficiency, but also damaging performance by introducing noises and diverting attention.
As such, we propose that by eliminating the sparsity of visual inputs, efficiency and performance can be optimized simultaneously, breaking the efficiency-performance trade-off in VLA models. More specifically, we investigate the underlying sparsity of visual tokens in VLA models, and propose LightVLA, a performance-driven differentiable visual token pruning framework for efficient VLA.
To evaluate the importance of visual tokens for task execution, LightVLA generates dynamic queries through the cross attention between visual tokens and task instruction tokens. Subsequently, each query selects a useful token in a differentiable manner using the Gumbel-softmax~\cite{gumbel} technique. Following the fine-tuning paradigm, we take OpenVLA-OFT~\cite{oft} as the foundation model and train LightVLA to distinguish and retain only informative visual tokens that contribute to overall performance. As shown in Fig.~\ref{fig:light_vit_1}, LightVLA obtains state-of-the-art performance on the LIBERO benchmark with significantly fewer visual tokens. Compared to OpenVLA-OFT, LightVLA achieves a 59.1\% reduction in total FLOPs and 2.6\% improvement in task success rate, highlighting that efficiency and performance are collateral goals that can be achieved simultaneously. 
To further fill in the existing gap in visual token pruning for VLA models, we propose LightVLA$^*$ in Discussion (Section \ref{sec:discussion}), which is an efficient and effective variant of LightVLA which introduces learnable query as additional trainable parameters to guide the model to select informative tokens.
To sum up, our contributions are outlined as follows:
\begin{itemize}[leftmargin=0.3cm]
    \item We empirically show that performance and efficiency can be collaterally optimized for VLA models.
    \item We propose LightVLA, a performance-driven differentiable visual token pruning framework for VLA models.
    \item Comprehensive experiments on the LIBERO benchmark demonstrate the state-of-the-art performance and efficiency of LightVLA compared to its foundation model and other previous models.
    \item To further fill in current gap in token pruning for VLA models, we propose LightVLA$^*$ an early exploration on learnable query-based token pruning, which also improves the performance and efficiency.
\end{itemize}

\section{Related Work}
\label{sec:relatedWork}
\textbf{Vision-language model (VLM).} 
Vision-language models (VLMs) integrate vision and language modalities, extending the reasoning capabilities of large language models (LLMs) to process visual input. 
This integration is typically achieved by encoding images into hundreds of visual tokens aligned with text tokens \cite{llava}.
Representative VLMs include LLaVa~\cite{llava}, BLIP-2~\cite{blip}, InternVL~\cite{internvl}, and Qwen-VL~\cite{qwenvl}, whose parameter spaces typically range from 7B to 70B. 
Despite these strengths, VLMs are not inherently designed to directly generate task-specific robotic action policies, which leads to the birth and emergence of VLA models.

\textbf{Vision-language-action (VLA) model.} 
VLAs \cite{openvla,pi0,pifast,gr-3} extend VLMs for embodied intelligence to generate feasible action policies for complicated robotic tasks like manipulating, bridging the gap between perception and action.
Similar to VLMs, representative VLA models (e.g. OpenVLA~\cite{openvla}, $\pi_0$~\cite{pi0} and CogACT~\cite{cogact}) typically tokenize image patches into hundreds of visual tokens, and then concatenate them with task tokens, from which actions are generated either as discrete tokens~\cite{openvla,worldvla,nora} or continuous values~\cite{cogact,oft}.
Early VLA models like OpenVLA generate discrete action tokens to further yield the action policy in the auto-regressive way, while recent VLA models aim to generate continuous action tokens like CogACT and OpenVLA-OFT.  Action chunking technique \cite{nora,oft,smolvla}, which aims to predict a sequence of action policies, has also demonstrated potential to improve the overall performance.
However, the billion-scale parameter and high inference cost of VLA models make them challenging to deploy in real-time low-latency robotic tasks.
To optimize the computation overhead and latency, existing works often aim to design lightweight VLA models, such as TinyVLA~\cite{tinyvla}, SmolVLA~\cite{smolvla}, and NORA~\cite{nora}. 
Beyond model architectures, token pruning offers a promising direction for optimizing efficiency with fewer input tokens, yet remains underexplored in VLA studies.

\begin{figure*}[htbp]
    \centering
    \includegraphics[width=1\linewidth]{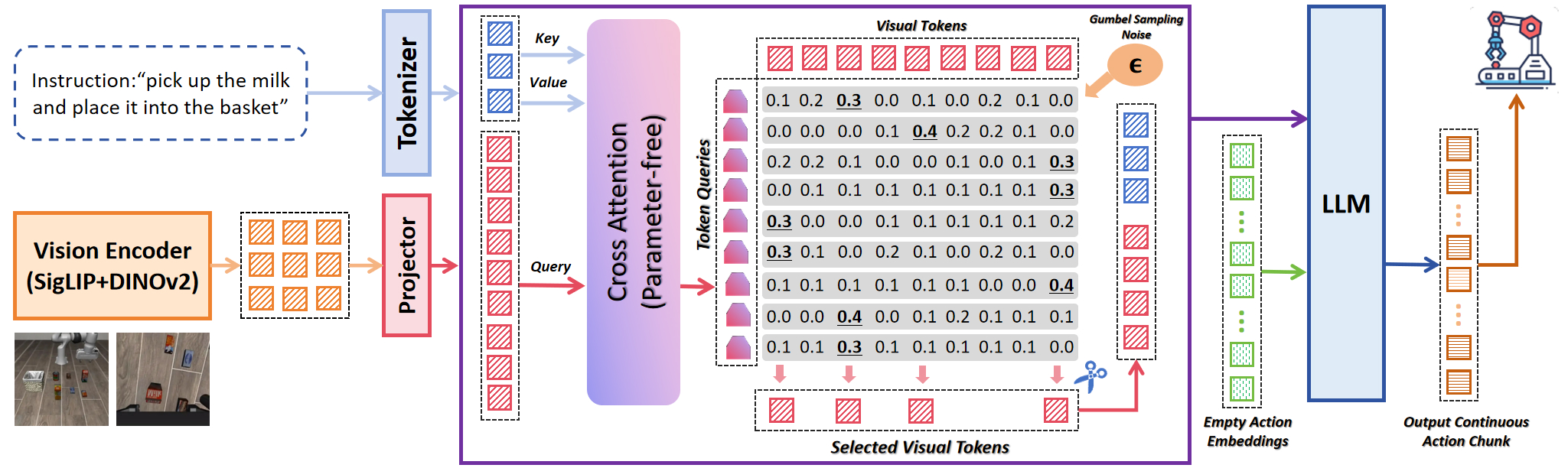}
    \caption{Illustration of the proposed LightVLA framework. Gray regions indicate the use of Gumbel-softmax for differentiable token selection.}
    \label{fig:framework}
\end{figure*}

\textbf{Visual token pruning.}
Token pruning has been successfully applied to diverse  neural architectures ranging from the original transformer and ViT~\cite{tome,ltp,fastvggt} to modern large-scale models like LLMs and VLMs~\cite{llavamini,fastv,sparsevlm,prumerge,pyramiddrop,lvpruning,VisPruner,CDPruner}.
Existing methods often pre-define a hyperparameter which determines a fixed number of retained visual tokens, necessitating large empirical exploration to select the optimal value of this hyperparameter.
In the context of VLA models, visual token pruning methods optimized specifically for VLMs tend to underperform when extended to VLA models, as analyzed in previous works \cite{spvla,vlacache,flashvla}. Existing works \cite{flashvla,spvla,vlacache, efficientvla} dedicated to VLA explore the potential of training-free visual token pruning with the guidance like the attention scores, which still relies on a fixed token pruning ratio.
This compression ratio present two major limitations: it introduces a strong inductive bias, potentially limiting the model's adaptability to varying visual inputs and tasks, and this approach often faces an inevitable performance drop.
Different from existing efforts, our proposed differentiable visual token pruning framework LightVLA dynamically selects necessary tokens for each task scenario, leading to improved performance while maintaining significant computational efficiency.
On the other hand, some well-noted inference-oriented platforms like vLLM~\cite{vllm} and SGLang~\cite{sglang} are optimized for high-throughput generation and they do not expose intermediate attention scores during inference. 
This makes traditional attention score-based token selection methods \cite{efficientvla,fastv} infeasible.
LightVLA does not require the attention scores from from within LLM, which can be well supported by these platforms, facilitating practical real-world deployment.

\section{Method}\label{sec:method}

We introduce LightVLA, a differentiable visual token pruning framework for VLA acceleration.
Unlike previous studies, our strategy is purely performance-driven, meaning that the only optimization goal is better performance. LightVLA enables the model to adaptively retain and discard tokens, during which time the model spontaneously learns to retain only useful tokens to maximize performance, thereby improving efficiency. Moreover, LightVLA requires no extra parameters, hyper-parameters or auxiliary losses, making it a universal framework compatible with most VLAs.

\subsection{Problem Definition}

A typical VLA can be decomposed into three components: A visual encoder with a projector $f_{v}$, a LLM backbone $f_{\phi}$, and an action head (or de-tokenizer) $f_{a}$. The visual encoder encodes the input image or images $X_{I}$ into $L_{v}$ initial visual tokens $H_{v^{\scriptscriptstyle\prime}}=f_{v}(X_{I}) \in \mathbb{R}^{L_{v}\times D^{\scriptscriptstyle\prime}}$. The initial visual tokens are  projected to $H_{v}=f_{v}(X_{I}) \in \mathbb{R}^{L_{v}\times D}$ and then concatenated with language tokens $H_{l} \in \mathbb{R}^{{L_l}\times D}$ and sent into the LLM backbone. Herein, $D^\prime$ and $D$ denote the initial vision token embedding dimension and the text token embedding dimension, respectively. The action head $f_{a}$ finally translates the output hidden states from $f_{\phi}$ into action policies oriented for robotic tasks.

The LLM outputs the final hidden states (i.e. output tokens) $H=f_{\phi}(H_{v},H_{l})$.
The computational bottleneck mainly lies in the decoder layers of the LLM $f_{\phi}$, which involves extensive attention-based computation of all visual tokens $H_{v}$ and text tokens $H_{l}$.  
Eventually, the action head converts $H$ of LLM into continuous action policies $A=f_{a}(H)$. 
Since $L_{v} >> L_{l}$ in most cases, one effective method to decrease the computation overhead is to introduce an efficient visual token pruner $f_{p}$.
The objective of a specific visual token pruner $f_{p}$ is to determine the pruned set of visual tokens $H'_{v} = f_{p}(H_{v}) \subseteq H_{v}$ to be kept, in order to reduce the computational cost of VLA without compromising its performance.
Note that we keep the $[\mathrm{CLS}]$ token since it maintains the global visual information \cite{vit}, and conduct pruning only on the patch-level visual tokens.

\subsection{LightVLA}
\label{sec:lightvla}

In previous studies~\cite{llavamini,fastv,sparsevlm,prumerge,pyramiddrop,lvpruning,VisPruner,CDPruner}, visual token pruners usually reduce the number of visual tokens to a pre-defined value $L'_{v}$. While simple and effective, this practice also comes with the risk of performance degradation due to information loss, especially when the task and scenario get complicated that $L'_{v}$ visual tokens cannot convey sufficient information. As such, it is imperative for VLA models to dynamically determine $H'_{v}$ according to its inputs so that the information loss can be further minimized.

In this study, we propose LightVLA, a novel query-based visual token pruning strategy that can adaptively distinguish informative visual tokens from $H_{v}$. As shown in Fig.~\ref{fig:framework}, LightVLA employs a series of $L_v$ token queries $Q=\{q_1,q_2,\cdots,q_{L_v}\}$,
each responsible for selecting one useful visual token from all tokens $h_k=q_i(H_v)$. All visual tokens selected by the queries constitute the pruned set $H'_v=\{\,h_k \mid \exists\, q_i,\; h_k=q_i(H_v)\,\}$. In the extreme condition when each query chooses a unique token, all tokens are selected, thus $H'_v = H_v$. When multiple queries choose the same token, repeated tokens will not be selected, and therefore $H'_v \subset H_v$. The pruning process can be divided into three steps: Query generation, token scoring, and token selection.

\subsubsection{\textbf{Query Generation}}
\label{sec:queryG}
While it is a common practice in prior VLM works to adopt learnable embeddings as queries \cite{blip,llavamini}, it also introduces extra parameters into the model, rendering this approach less feasible in resource-constrained platforms that VLA is often implemented on.
As such, we propose a parameter-free query generation method for better compatibility.

We note that the usefulness of a visual token can be reflected by the interactions between its visual information and the language instruction. For instance, when we give an instruction \emph{pick up the milk and place it into the basket} as the text prompt, the VLA model should focus more on the two semantic objects (i.e. milk and the basket) in the image instead of other less informative objects or background.

Therefore, queries can be generated via cross attention between visual tokens and language tokens.
\begin{equation}
\label{eq:queryGene}
    Q = \text{softmax}(\frac{H_vH_l^T}{\sqrt{D}})H_l,
\end{equation}
where $Q \in \mathbb{R}^{L_{v}\times D}$.
Note that unlike traditional attention design, neither weight nor bias matrices is included in the query generation process for simplicity.

\subsubsection{\textbf{Token Scoring}}

In this step, each query estimates the usefulness of all tokens individually through token scoring.
\begin{equation}
    S = \frac{QH_v^T}{\sqrt{D}}
\end{equation}
Here, $S \in \mathbb{R}^{L_v \times L_v}$ is the score matrix, where each element $s_{i,j}$ denotes the score of the $j$-th token assigned by the $i$-th query.

\subsubsection{\textbf{Token Selection}}

To determine the pruned token set, each query selects the token with the highest score.
\begin{equation}
    H'_v=\{h_k | k=\text{argmax}_j(s_{i,j}), j=1,2,\cdots,L_v\}.
\end{equation}

However, during training, it is noted that the $\text{argmax}$ operation is not differentiable. One solution proposed in previous studies~\cite{lvpruning} is to incorporate auxiliary loss on the token scores.
However, the introduction of auxiliary loss not only complicates the training procedures, but may also lead to performance deterioration and gradient conflicts of different optimization objectives, and the ground truth of token scores is difficult to define.

To overcome this obstacle, we adopt the Gumbel-softmax sampling technique~\cite{gumbel} to make the $\text{argmax}$ operation differentiable. This technique allows the process of sampling from discrete tokens to be differentiable in the backward process, which guides the VLA model to learn to select most informative visual tokens.
Specifically, we convert the score matrix $S \in \mathbb{R}^{L_v \times L_v}$ into an indicator matrix $I \in \mathbb{R}^{L_v \times L_v}$.
\begin{align}
    S' &= S+\epsilon \\
    S_\text{soft} &= \text{softmax}_j(S') \\
    S_\text{hard} &= \text{one-hot}(\text{argmax}_j(S')) \\
    I &= S_\text{hard} + S_\text{soft} - S_\text{soft}^{\textit{SG}}
\end{align}
where $S'$ is the score matrix injected with Gumbel sampling noise $\epsilon \in U(0, \alpha)$, $S_\text{soft}$ and $S_\text{hard}$ are the soft and hard scores, $\text{one-hot}$ is the one-hot function, and $SG$ indicates the stop gradient operation. 

Note that unlike the original Gumbel-softmax operation where $\epsilon \in U(0,1)$, we gradually decrease the intensity level of sampling noise as training progresses by decaying the noise upper bound $\alpha$. 
We also provide corresponding ablation study in Section \ref{sec:abla}.
This design encourages model to better explore more diverse token selection schemes in the early learning stage, and helps the model stabilize in the final stage.
With the indicator matrix $I$, the pruned set can be obtained by
\begin{equation}
    H'_v = I H_v^T
\end{equation}
Here, since $I$ is an indicator matrix, $H'_v$ only contains the tokens selected by the queries.
Moreover, the gradient of $I$ equals to the gradient of $S'$. Thus, the queries can be correspondingly optimized in an end-to-end manner with the gradient descent.
Notably, for inference, we follow the direct argmax operation to pick up visual tokens selected by queries without the Gumbel noise.

Moreover, we notice that the LLM backbone relies heavily on the position IDs of visual tokens to understand the spatial relationship. Therefore, the position IDs are retained during the token selection process.

\section{Experiments}
\label{sec:exp}
\subsection{Experimental Settings}
\textbf{Dataset.} We evaluate LightVLA on the LIBERO benchmark \cite{libero}, which features a Franka Emika Panda arm in simulation with demonstrations containing camera images, robot state, task annotations, and delta end-effector pose actions. 
We use four task suites including LIBERO-Spatial, LIBERO-Object, LIBERO-Goal, and LIBERO-Long, each of which provides 500 expert demonstrations across 10 tasks. We test LightVLA on all tasks for 50 trials (i.e. 500 trials in total for each suite) to assess policy generalization to different spatial layouts, object selection, task goals, and long-horizon planning tasks. We report the success rate (\%) on each task suite.

\textbf{Baselines.} 
For comparison, we compare the proposed LightVLA with  baselines of two types, where the first type is diverse VLA model design and the second type considers applying token pruning on existing VLA models:
\begin{itemize}[leftmargin=0.3cm]
    \item \textit{Diverse VLA models}: 
    OpenVLA \cite{openvla}, $\pi_0$ series \cite{pi0,pifast}, NORA series~\cite{nora}, SmolVLA~\cite{smolvla}, OpenVLA-OFT~\cite{oft}, CogACT \cite{cogact}. WorldVLA with 256 visual tokens and WorldVLA* with 1024 visual tokens \cite{worldvla}.
    \item \textit{Token pruning methods}: 
    FlashVLA~\cite{flashvla}, SP-VLA~\cite{spvla}, VLA-cache\cite{vlacache}, FastV \cite{fastv}, SparseVLM~\cite{sparsevlm}.
\end{itemize}

Note that SparseVLM and FastV are token pruning methods which are firstly proposed for VLMs yet used on the OpenVLA backbone following \cite{vlacache}.

\begin{table*}[t]
\caption{Experimental results on LIBERO benchmark. TP denotes token pruning. AR, FM and PD denote different action decoding (i.e. generation) paradigms including auto-regressive, flow matching, and parallel decoding. The average number and standard deviation of retained visual tokens of LightVLA are denoted under the success rate, respectively. $^*$ Our reproduced results, slightly different from the original paper~\cite{oft} due to hardware discrepancy.}
\label{tab:main}
\centering
\begin{adjustbox}{max width=\linewidth}
\begin{tabular}{l|ccc|c|cccc|c}
\toprule \toprule
\textbf{Method}  & \textbf{Scale} & \textbf{TP} & \textbf{Decoding} & \textbf{Backbone} & \textbf{Spatial SR}(\%) & \textbf{Object SR}(\%) & \textbf{Goal SR}(\%) & \textbf{Long SR}(\%) & \textbf{Avg}. \\ 
\midrule
OpenVLA \cite{openvla}  &7B & & AR & PrismaticVLM& 84.7 & 88.4 & 79.2 & 53.7 & 76.5 \\
SparseVLM \cite{sparsevlm} &7B & $\checkmark$ & AR & PrismaticVLM& 79.8 & 67.0 & 72.6 & 39.4 &  64.7 \\
FastV \cite{fastv} &7B & $\checkmark$ & AR & PrismaticVLM& 83.4 & 84.0 & 74.2 & 51.6 &  73.3 \\
VLA-Cache \cite{vlacache} &7B & $\checkmark$ & AR & PrismaticVLM& 83.8 & 85.8 & 76.4 & 52.8 &  74.7 \\
FlashVLA \cite{flashvla}  &7B  &$\checkmark$ & AR & PrismaticVLM& 84.2 & 86.4 & 75.4 & 51.4 & 74.4 \\
SP-VLA \cite{spvla}  &7B & $\checkmark$ & AR & PrismaticVLM& 75.4 & 85.6 & 84.4 & 54.2 & 74.9 \\
WorldVLA \cite{worldvla} & 7B && AR & Chameleon & 85.6 &89.0 &82.6 &59.0 &79.1 \\
WorldVLA* \cite{worldvla} & 7B && AR & Chameleon & 87.6 &96.2 &83.4& 60.0 &81.8 \\
NORA \cite{nora} & 3B  && AR & Qwen-VL & 85.6 & 87.8 & 77.0 & 45.0 & 73.9 \\
SmolVLA \cite{smolvla}  &2.25B && FM & SmolVLM & 93.0 & 94.0 & 91.0 & 77.0 & 88.8 \\
CogACT \cite{cogact} & 7B  && FM & PrismaticVLM& 97.2 & 98.0 & 90.2 & 88.8 & 93.6 \\ 
$\pi_0$ \cite{pi0}  &3.3B && FM & PaliGemma & 96.8 & \textbf{98.8} & 95.8 & 85.2 & 94.2 \\
$\pi_0$-fast \cite{pifast}  &3.3B && FM & PaliGemma & 96.4 & 96.8 & 88.6 & 60.2 & 85.5 \\
NORA-Long \cite{nora} & 3B  && PD & Qwen-VL & 92.2 & 95.4 & 89.4 & 74.6 & 87.9 \\
OpenVLA-OFT$^*$ \cite{oft}  &7B && PD & PrismaticVLM& 97.6 & 94.2 & 95.2 & 92.0 & 94.8 \\
\midrule
LightVLA (Ours)  &7B &$\checkmark$& PD & PrismaticVLM& \begin{tabular}[c]{@{}c@{}}\textbf{98.4}\\ (90$\pm$15 tokens)\end{tabular} & \begin{tabular}[c]{@{}c@{}}98.4\\ (78$\pm$11 tokens)\end{tabular} & \begin{tabular}[c]{@{}c@{}}\textbf{98.2}\\ (64$\pm$10 tokens)\end{tabular} & \begin{tabular}[c]{@{}c@{}}\textbf{94.6}\\ (79$\pm$11 tokens)\end{tabular} & \textbf{97.4} \\
\bottomrule
\end{tabular}
\end{adjustbox}
\vspace{0.5em}
\end{table*}

\textbf{Implementation details.}
All experiments are conducted on 8 Nvidia$\circledR$ H20 GPUs.
We use the open-sourced OpenVLA-OFT \cite{oft} as the foundation model which consists of the two-branch vision encoder (DINOv2 \cite{dinov2} and SigLIP \cite{siglip}), LLaMA-2-7B \cite{llama} as the language model backbone. 
The LoRA \cite{lora} technique with rank $32$ is applied on the entire model, including the vision encoder, LLM backbone and action head, for fine-tuning. 
We initialize the foundation model with the open-sourced OpenVLA-OFT checkpoints on HuggingFace, and then the model is fine-tuned for 40,000 gradient steps in total and we decay the learning rate from $5e-4$ to $5e-5$ after 30,000 gradient steps. 
The batch size on each device is $8$, resulting in a global batch size of $64$.

\subsection{Main Experiments}
We evaluate our proposed LightVLA against other baselines constructed by diverse foundation model architectures on the LIBERO benchmark in Table \ref{tab:main}.
The results of CogACT are obtained from \cite{geovla}.
As the results indicate, LightVLA delivers the best performance across all tasks. Notably, LightVLA outperforms its baseline model, OpenVLA-OFT on all task suites by a large margin. Moreover, compared to OpenVLA-OFT which consumes 512 visual tokens, LightVLA only retains 78 visual tokens on average, indicating that most visual tokens are not contributing to overall performance. The results not only shed light on the sparsity of visual modality, but also proves that performance and efficiency are collateral goals that can be optimized simultaneously.

\subsection{Analysis on Computation Efficiency}

Here, we present the comparison of acceleration and performance of LightVLA and previous VLA acceleration approaches in Table~\ref{tab:acceleration}. Results show that LightVLA achieves the highest average success rate while substantially reducing latency and computational demands compared with other strong baselines. Compared to OpenVLA-OFT, LightVLA not only reduces FLOPs and latency by 59.1\% and 38.2\%, but also improves success rate by 2.6\%. Remarkably, among all VLA acceleration approaches in Table~\ref{tab:acceleration}, LightVLA is the only one that boosts performance. Our findings prove that in the pursuit of optimal performance, the efficiency of VLA models can also be optimized due to the elimination of visual sparsity.

\renewcommand{\arraystretch}{1.2}
\begin{table}[htbp]
\caption{Comparison of acceleration and performance on the LIBERO benchmark between LightVLA and other methods. We report visual token counts, GPU types, computational cost (TFLOPs), end-to-end latency, and averaged task success rate.\label{tab:acceleration}}
\centering
\setlength{\tabcolsep}{4pt}
\begin{adjustbox}{max width=\linewidth}
\begin{tabular}{l l c c c c}
\toprule\toprule
\textbf{Method} & \textbf{Visual Token} & \textbf{GPU} & \textbf{TFLOPs} & \textbf{Latency (ms)} & \textbf{SR (\%) Avg.} \\
\midrule
OpenVLA & 256 & A100 & - & - & 76.5 \\
SparseVLM & 100 (max.) & RTX 4090 & 1.4 & 83  & 64.7 \\
FastV & 100 (max.) & RTX 4090 & 1.9 & 53  & 73.3 \\
VLA-Cache & 100 (max.) & RTX 4090 & 1.4 & 32  & 74.7 \\
FlashVLA & 192 & H100 & 0.7 & 55 & 73.7 \\
SP-VLA & 229 (avg.) & A100 & 3.1 & - & 74.9 \\ 
\midrule
OpenVLA-OFT & 512 & H20 & 8.8 & 34 & 94.8 \\
LightVLA & 78 (avg.) & H20 & 3.6 & 21 & 97.4 \\
\bottomrule
\end{tabular}
\end{adjustbox}
\end{table}

\subsection{Ablation Study}
\label{sec:abla}

\begin{figure*}[htbp]
    \centering
    \includegraphics[width=1\linewidth]{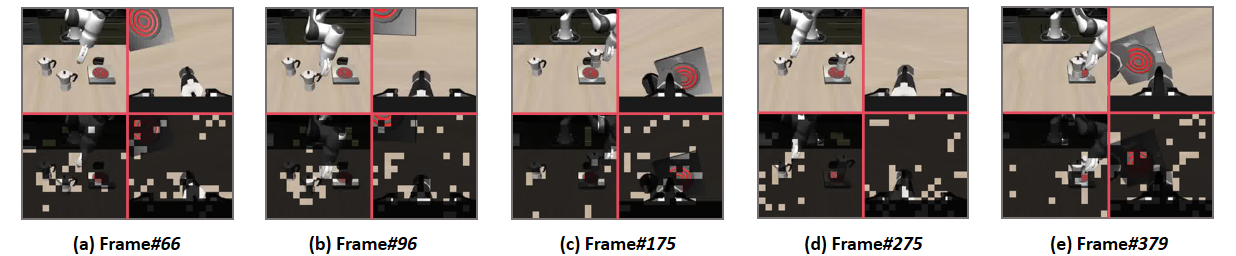}
    \caption{An example of LIBERO-Long task: 'Put both moka pots on the stove'. Each frame consists of 4 images. Upper left: The 3rd person view camera. Upper right: The wrist camera. Lower left: The 3rd person view camera with pruned tokens masked. Lower right: The wrist camera with pruned tokens masked.}
    \label{fig:frames_vis}
\end{figure*}

\textbf{Impact of noise factor schedule.} In this study, we propose to gradually decrease the intensity level of sampling noise for more diverse token selection schemes. To validate this technique, we compare LightVLA with two variants: (1) LightVLA without sampling noise, and (2) LightVLA with constant sampling noise (i.e. without noise decay). The results are presented in Table~\ref{tab:noise-factor}, which show that LightVLA outperforms both variants. A deeper analysis into the token pruning schemes of these variants reveals why noise factor schedule works. Compared to LightVLA, the variant without sampling noise retains fewer visual tokens, which oftentimes leads to loss of important semantic information, especially in semantically dense scenarios such as Object and Goal. The introduction of sampling noise  alleviates this problem by encouraging more diverse token pruning choices. Besides, the second variant shows that with constant sampling noise, LightVLA finds it hard to learn to prune tokens, resulting in a significantly higher token number.

\begin{table}[htbp]
\centering
\caption{Impact of noise factor schedule on LightVLA.}
\label{tab:noise-factor}
\begin{adjustbox}{max width=\linewidth}
\begin{tabular}{lcccccc}
\toprule \toprule
\textbf{Variant} & \begin{tabular}[c]{@{}c@{}}Spatial\\ SR(\%)\end{tabular} & \begin{tabular}[c]{@{}c@{}}Object\\ SR(\%)\end{tabular} & \begin{tabular}[c]{@{}c@{}}Goal\\ SR(\%)\end{tabular} & \begin{tabular}[c]{@{}c@{}}Long\\ SR(\%)\end{tabular} & \textbf{Avg}. & \begin{tabular}[c]{@{}c@{}}\# Tokens\\ (Avg.)\end{tabular} \\ \midrule
LightVLA & 98.4 & \textbf{98.4} & \textbf{98.2} & 94.6 & \textbf{97.4} & 78 \\
- w/o noise  & 98.8 & 97.6  & 97.2 & 94.2 & 97.0 & 72 \\
- w/o schedule  & \textbf{99.4} & 97.8  & 96.0 & \textbf{94.8} & 97.0 & 112 \\\bottomrule
\end{tabular}
\end{adjustbox}
\end{table}

\textbf{Impact of retained tokens on performance.}
One highlight of LightVLA is its ability to adaptively distinguish useful tokens. To validate this ability, we manipulate the token pruning scheme of LightVLA in the following ways. First, after LightVLA has retained $k$ tokens, we supplement another $k$ random tokens into the pruned set, resulting in $2k$ tokens sent into the LLM. This manipulation validates if any useful tokens have been omitted by LightVLA. Second, after LightVLA has retained $k$ tokens, we randomly discard $10\%$ of the tokens from the pruned set, resulting in $0.9k$ tokens sent into the LLM. This manipulation tests if any useless tokens have been retained by LightVLA. The results shown in Table~\ref{tab:tokens} demonstrate that any manipulation to the token pruning scheme will result in performance degradation, proving LightVLA's ability to retain only useful tokens and discard useless tokens.

\begin{table}[htbp]
\centering
\caption{Impact of retained tokens on performance.}
\label{tab:tokens}
\begin{adjustbox}{max width=\linewidth}
\begin{tabular}{lcccccc}
\toprule \toprule
\textbf{Model} & \# Tokens & \begin{tabular}[c]{@{}c@{}}Spatial\\ SR(\%)\end{tabular} & \begin{tabular}[c]{@{}c@{}}Object\\ SR(\%)\end{tabular} & \begin{tabular}[c]{@{}c@{}}Goal\\ SR(\%)\end{tabular} & \begin{tabular}[c]{@{}c@{}}Long\\ SR(\%)\end{tabular} & \textbf{Avg}. \\ \midrule
LightVLA & $k$ & \textbf{98.4} & \textbf{98.4} & \textbf{98.2} & \textbf{94.6} & \textbf{97.4} \\
LightVLA & $2k$ &  98.0 & 97.6  & 97.8 & 93.8 & 96.8 \\
LightVLA & $0.9k$ &  98.2 & 97.8  & 97.2 & 93.0 & 96.6 \\\bottomrule
\end{tabular}
\end{adjustbox}
\end{table}

\subsection{Qualitative Visualization}
To better illustrate the token pruning process, we take an episode as a demonstration to show the token selection dynamics when asking the VLA model to perform the manipulation task. The key frames are selected to represent critical phases of the manipulation task, including object interaction, and task completion. As shown in Fig.~\ref{fig:frames_vis}, the retained tokens concentrate on objects of interest, i.e. moka pots, stove, and the robotic arm itself, whereas most background tokens are pruned. 
Besides, it also shows that LightVLA learns to retain or prune more tokens when needed, as exemplified by the comparison between Frame\textit{$\#$175} and Frame\textit{$\#$275}.
The visualization results further demonstrate the effectiveness of LightVLA in adaptive token pruning.


\section{Discussion}
\label{sec:discussion}

\subsection{Learnable Query for Token Pruning}

Since token pruning (especially the training-aware token pruning approach) is rarely investigated in the VLA research, besides the previously introduced parameter-free token pruning method LightVLA, we also implement LightVLA$^*$, which exploits the \textit{learnable query} with extra trainable parameters \cite{llavamini,blip} to select informative visual tokens. 
We consider applying this learnable query on two different positions which considers only the visual features or the joint visual-language features.
Specifically, to filter out redundant visual tokens, we introduce $N_q$ compression queries as token query head, designed to guide the model to learn to select visual tokens from all $L_v$ visual tokens. 

\begin{figure}[htbp]
    \centering
    \includegraphics[width=1\linewidth]{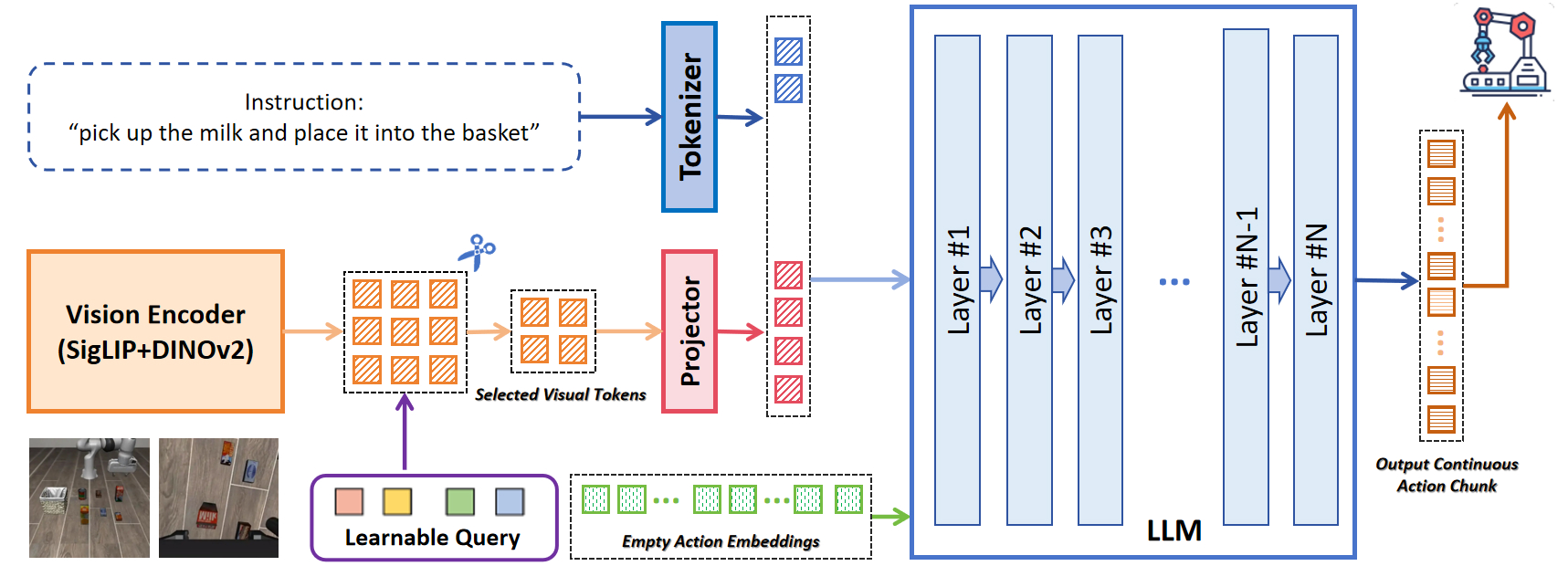}
    \caption{Illustration of LightVLA$^*$ when pruning visual tokens at the vision encoder with the learnable query.}
    \label{fig:light_vit}
\end{figure}

\textbf{Learn to select at the vision encoder.}
The $N_q$ compression queries interact with all visual tokens $H_v$, selectively extracting the important visual information to produce the selected visual tokens. Different from Eq. \ref{eq:queryGene} of the parameter-free LightVLA, we introduce the learnable query  $Q^* \in \mathbb{R}^{N_q \times D^\prime}$ after the vision encoder, as visualized in Figure \ref{fig:light_vit}. Note that visual tokens are firstly concated at the channel level and then we perform our token pruning. Then we compute the token scoring as follows:

\begin{equation}
    S^* = \frac{LN(Q^*)\cdot LN(H_v^T)}{\sqrt{D^\prime}},
\end{equation}
where $D^\prime$ denotes the dimension of visual tokens before the projection layer, $LN$ denotes the layer normalization layer to stablize the training process and $S^* \in \mathbb{R}^{N_q\times L_v}$ denotes the score matrix. 
We utilize the RMSNorm \cite{rms} to achieve the efficient layer normalization.
According to $S^*$, each query selects the visual token with the highest score and the overall training process is differentiable via the Gumbel-softmax operation, which is consistent with LightVLA. LightVLA$^*$ introduces the learnable query $Q^*$ and the mapping parameters of layer normalization as the extra parameters, and it can be supported by the efficient inference-oriented platforms like vLLM and SGLang.

\begin{figure}[htbp]
    \centering
    \includegraphics[width=1\linewidth]{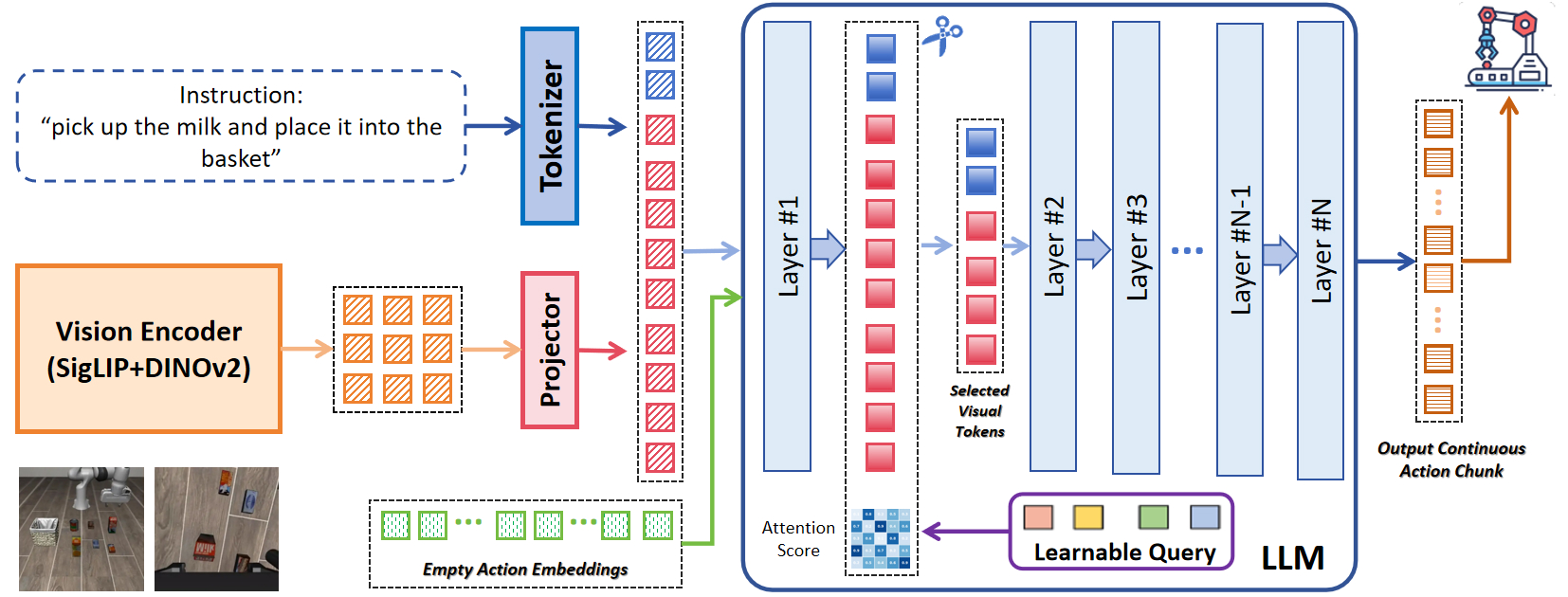}
    \caption{Illustration of LightVLA$^*$ when pruning visual tokens at the first decoder layer of LLM with the learnable query.}
    \label{fig:light_layer1}
\end{figure}

\textbf{Learn to select at the early layer of LLM.}
Instead of operating at the image encoder, the compression queries $Q_v$ can interact with both visual tokens $H_v$ and early-layer text tokens $H_l$ in the LLM, as visualized in Figure \ref{fig:light_layer1}.
This allows the model leverages the semantic information from the task-level text to guide the token pruning process. The main idea is similar to the above part 'Learn to select at the vision encoder', but we introduce the attention scores (text tokens to visual tokens) as follows:
\begin{equation}
    S^\dagger = \frac{LN(Q^\dagger)\cdot LN(H_v^T)+\zeta \cdot attn}{\sqrt{D}},
\end{equation}
where $D$ denotes the projected vision token dimension, $attn$ indicates the attention scores of visual tokens to the text tokens, $Q^\dagger \in\mathbb{R}^{N_q \times D}$ denotes the learnable query and $\zeta$ denotes the learnable trade-off weight of the cross attention weights which is initialized with $1.0$. The attention score is output by the corresponding decoder layer of LLM.
We prune redundant visual tokens at early layers (layer\#1 to layer\#3) instead of deeper layers (layer\#4 to layer\#32), as the deep layers have already developed rich cross-modal representations where visual and textual features are deeply fused and semantically entangled.
Another reason lies in pruning visual tokens at early layers of LLM facilitates the better computation decrease since each decoder layer requires extensive quadratic attention-based computation.
Note that when pruning visual tokens at the decoder layer, LightVLA$^*$ requires output attention scores which can not be supported by above mentioned inference-oriented platforms.

\textit{\textbf{Analysis}.} In LightVLA$^*$ experiments, we initialize $N_q=128$ queries, which firstly retains 25\% of visual tokens since our LightVLA indicates adaptively maintaining 10\% to 20\% visual tokens is sufficient. 
Experimental results are listed in Table \ref{tab:learnable}. We find both LightVLA$^*$ series and LightVLA achieves better performance against their counterparts in Table \ref{tab:main}. LightVLA$^*$ at the first decoder layer can achieve the best performance for the complex long-horizon LIBERO-long task suite. We can also observe a slight averaged performance decrease when pruning visual tokens in the deeper decoder layer.

\begin{table}[htbp]
\centering
\caption{Experimental results with learnable token query. SR denotes the success rate (SR).}
\label{tab:learnable}
\begin{adjustbox}{max width=\linewidth}
\begin{tabular}{lcccccc}
\toprule \toprule
\textbf{Method}  &\begin{tabular}[c]{@{}c@{}}\textbf{Pruning}\\ \textbf{Position}\end{tabular} & \begin{tabular}[c]{@{}c@{}}\textbf{Spatial}\\ \textbf{SR}(\%)\end{tabular} & \begin{tabular}[c]{@{}c@{}}\textbf{Object}\\ \textbf{SR}(\%)\end{tabular} & \begin{tabular}[c]{@{}c@{}}\textbf{Goal}\\ \textbf{SR}(\%)\end{tabular} & \begin{tabular}[c]{@{}c@{}}\textbf{Long}\\ \textbf{SR}(\%)\end{tabular} & \textbf{Avg}. \\ \midrule
OpenVLA-OFT  & - & 97.6 & 94.2 & 95.2 & 92.0 & 94.8 \\ \midrule
LightVLA$^*$  & Vision Encoder & 98.2 & 96.2  & 96.2 & 94.2 & 96.2 \\
LightVLA$^*$  &LLM (layer\#1)& 98.0 & \textbf{98.0} & 97.2 & \textbf{94.8} & 97.0  \\
LightVLA$^*$  &LLM (layer\#2)& 97.8 & \textbf{98.0} & 96.6 & 94.2 & 96.7 \\
LightVLA$^*$  &LLM (layer\#3)& 97.6 & \textbf{98.0} & 97.0 & 93.8 & 96.6  \\ \midrule
LightVLA  & Vision Encoder & \textbf{98.4} & 97.6 & \textbf{98.2} & 94.6 & \textbf{97.4} \\ \bottomrule
\end{tabular}
\end{adjustbox}
\end{table}

\subsection{Comparison with MoE}

It is obvious that LightVLA and the mixture of expert (MoE) technique \cite{moe} share the similar intuition: 
Both techniques select a dense subset of elements from the whole collection to optimize the forward efficiency.
The main dissimilarity between LightVLA and MoE is their different goals, which in turn lead to different behaviors. LightVLA aims to maximize performance while improving efficiency. Therefore, LightVLA focuses its token selection to only informative tokens.
In contrast, MoE is proposed to divide specialized tasks into subtasks handled by experts.
To balance knowledge and workload among experts, MoE distributes its selection evenly among experts without a particular focus. In conclusion, LightVLA and MoE are fundamentally different in both goals and behaviors, making them distinct techniques.

\section{Conclusion}
\label{sec:conclusion}
In this work, we investigate the inherent visual redundancy of vision-language-action (VLA) models, and propose the extra parameter-free visual token pruning framework LightVLA. 
With differentiable query-based token pruning process, it adaptively selects informative visual tokens.
It achieves the state-of-the-art performance on the LIBERO benchmark with the significant computational optimization. We also propose another framework LightVLA$^*$ with learnable query as additional trainable parameters, which also outperforms against its counterparts. 

Regarding future works, our research schedule is two-fold. Firstly, we aim to further investigate the visual redundancy in end-to-end VLMs or VLA models oriented for autonomous driving to optimize the overall efficiency and latency which facilitates the real-world deployment. Meanwhile, we plan to explore the efficient token pruning to VLMs or VLA models designed for complicated spatial intelligence tasks which facilitates the wider application of consumer-level devices like household robots.

\section*{ACKNOWLEDGMENT}
We thank Pengxiang Li from LiAuto for discussion and Fang Yang from Tsinghua University for writing advice.


\bibliographystyle{IEEEtran}
\bibliography{IEEEabrv,refs}

\end{document}